\documentclass[sigconf]{acmart}

\usepackage{subcaption}

\newtheorem{definition}{Definition}
\newtheorem{Conjecture}{Conjecture}

\AtBeginDocument{%
  }

\copyrightyear{2025}
\acmYear{2025}
\setcopyright{rightsretained}

\acmConference[WWW Companion '25]{Companion Proceedings of the ACM Web Conference 2025}{April 28-May 2, 2025}{Sydney, NSW, Australia}
\acmBooktitle{Companion Proceedings of the ACM Web Conference 2025 (WWW Companion '25), April 28-May 2, 2025, Sydney, NSW, Australia}
\acmDOI{10.1145/3701716.3715598}
\acmISBN{979-8-4007-1331-6/2025/04}

\makeatletter
\gdef\@copyrightpermission{
  \begin{minipage}{0.2\columnwidth}
  \end{minipage}\hfill
  \begin{minipage}{0.8\columnwidth}
   \href{https://creativecommons.org/licenses/by-nc-sa/4.0/}{This work is licensed under a Creative Commons Attribution-NonCommercial-ShareAlike International 4.0 License.}
  \end{minipage}
  \vspace{5pt}
}
\makeatother

\begin{document}

\title{Fairness and Robustness in Machine Unlearning}

\author{Khoa Tran}
\affiliation{%
  \institution{
  Department of Computer Science and Engineering \\ Sungkyunkwan University}
  \city{Suwon}
  \country{Republic of Korea}
}
\email{khoa.tr@g.skku.edu}

\author{Simon S. Woo}
\orcid{0000-0002-8983-1542}
\authornote{Corresponding author. Email: swoo@g.skku.edu (Simon S. Woo)}
\affiliation{%
  \institution{
  Department of Computer Science and Engineering \\ Sungkyunkwan University}
  \city{Suwon}
  \country{Republic of Korea}
}
\email{swoo@g.skku.edu}

\begin{abstract}
Machine unlearning poses the challenge of ``how to eliminate the influence of specific data from a pretrained model'' in regard to privacy concerns. While prior research on approximated unlearning has demonstrated accuracy and efficiency in time complexity, we claim that it falls short of achieving exact unlearning, and we are the first to focus on fairness and robustness in machine unlearning algorithms. Our study presents fairness Conjectures for a well-trained model, based on the variance-bias trade-off characteristic, and considers their relevance to robustness. Our Conjectures are supported by experiments conducted on the two most widely used model architectures—ResNet and ViT—demonstrating the correlation between fairness and robustness: \textit{the higher fairness-gap is, the more the model is sensitive and vulnerable}. In addition, our experiments demonstrate the vulnerability of current state-of-the-art approximated unlearning algorithms to adversarial attacks, where their unlearned models suffer a significant drop in accuracy compared to the exact-unlearned models. We claim that our fairness-gap measurement and robustness metric should be used to evaluate the unlearning algorithm. Furthermore, we demonstrate that unlearning in the intermediate and last layers is sufficient and cost-effective for time and memory complexity.
\end{abstract}





\maketitle

\section{Introduction}

Deep neural networks have demonstrated remarkable performance across multiple domains. The key elements of its success are the powerful hardware and large amounts of data for training. However, due to the large model and data sizes, the model's behavior remains incomprehensible and uncontrollable. The model may unintentionally memorize its training data, raising concerns over privacy risks. The presence of private-sensitive user information in the training data poses a considerable issue in governing access to personal data ownership on the web.

In fact, the General Data Protection Regulation (GDPR) \cite{GDPR} and AI Act \cite{artificialintelligenceact} established by the European Union, enacted a privacy policy aimed at providing users control over their personal data. It requires that companies have to remove users' private dor copyrighted web data from the database and AI model upon request. Furthermore, this data must be invalid to any privacy threats, since attackers may employ diverse methods to reconstruct it. 

In classification problems, there are existing unlearning methods aimed at removing the influence of certain data from a pretrained model \cite{Chen2023BoundaryUR, Fan2023SalUnEM, Kurmanji2023TowardsUM, Hayase2020SelectiveFO}, which we detail on in the next section. Generally, they used accuracy as a metric for evaluating the impact of a dataset on the model, where low accuracy on a dataset indicates that the model has effectively ablated the forget class. This study demonstrates that while existing unlearning methods effectively remove specific data in terms of accuracy, they still present potential risks of fairness and vulnerability. We summarize our major contributions as follows:
\begin{itemize}
    \item Leveraging the variance-bias trade-off principle and its relevance to the variance property of batch norm, we provide our Conjectures regarding the connection between fairness and robustness properties. We claim that these attributes are crucial for a well-trained model and protect it from security and privacy threats.
    \item Through unlearning and adversarial attack experiments, we demonstrate that empirical results clearly align with our Conjectures. And, this emphasizes that the existing unlearning approaches inadequately address fairness properties, resulting in weakened robustness. 
    
    \item Finally, we propose that robustness evaluation be a critical metric for machine unlearning, and claim that unlearning in the intermediate and last layers is sufficient and efficient in terms of time and memory complexity to achieve exact unlearning.
\end{itemize}

\section{Preliminary and Related Works}

\textbf{Variance-bias trade-off.}
Bias-variance trade-off is a fundamental principle for understanding the generalization of predictive learning models \cite{VATradeoff}. The bias is an error term that measures the mismatch between the model's prediction and ground truth distribution, and the variance measures the sensitivity of predictions and how it fluctuates when there is a small change in the input. In the scope of this paper, we focus on variance properties, where the model is considered sensitive and overfitting if the distribution of features exhibits high variance on a dataset.

\textbf{Normalization layers.}
Feature normalization has proved its advantages in stability and speed-up training \cite{Ioffe2015BatchNA, Ba2016LayerN}. While staking a huge amount of layers in deep neural networks leads to high-variance output and overfitting, this technique aims to prevent the distribution shift by normalization feature vectors to a unit Gaussian distribution. And, Batch norm \cite{Ioffe2015BatchNA} normalizes each feature within a batch of samples, and is widely used in popular convolutional neural network models such as ResNet \cite{He2015DeepRL}, VGG \cite{Simonyan2014VeryDC}, etc. While Layer norm \cite{Ba2016LayerN} normalizes all features within each sample, it is well-known to be used in transformer models \cite{Vaswani2017AttentionIA}. Due to the distribution shift between each layer in the deep model, the distribution of feature vectors from normalization modules can be considered stable for the analysis of feature properties.

\textbf{Adversarial attack.} We revisit adversarial attack in classification setting. Given a model $f_\theta$ is trained on a dataset $\{x_i,y_i\}^N$, where $x\in\mathcal{X}$ is input and $y\in\{1,2,...,C\}, C\in N$ is a categorized label, the goal of an adversarial attack is to produce an adversarial example $x^{'}\in\mathcal{X}$, such that $f_\theta(x^{'})\neq y$. In this paper, we adopt FGSM \cite{Goodfellow2014ExplainingAH}, the most popular adversarial attack method that creates adversarial noise by the direction of the sign of gradient at each pixel as follows:
\begin{equation}
    x^{'} = x + \eta \ \text{sign} (\nabla_x\mathcal{L}(f_\theta(x), y)),
\end{equation}
where $\eta$ is the magnitude of the perturbation.

 \textbf{Machine unlearning.}
Machine unlearning techniques aim to remove the impact of particular data from a pretrained model. In classification problems, it can be classified into two scenarios: instance-wise unlearning and class-wise unlearning. Instance-wise unlearning aims to remove a subset of randomly chosen data from the training dataset, whereas class-wise unlearning intends to remove the influence of an entire data class. In this paper, we focus on the unlearning problem in a class-wise context, where a selected class for removal is referred to as the forget set, while the other classes are called the retain set.

In addition, machine unlearning methods are primarily divided into two approaches: exact unlearning and approximated unlearning. In exact unlearning, we remove the forget set from the original training data and then retrain the model from the beginning. This strategy is not always feasible, especially when the training data is large, as retraining the entire model faces high computational costs and is time-consuming. Furthermore, data privacy concerns restrict access to the original training data in the commercial AI system. Consequently, approximated unlearning is proposed to enable more efficient unlearning. That is, starting with a trained model, approximated unlearning incrementally adjusts the model's weights over a constrained period of time. It results in significantly reduced costs relative to exact unlearning.

\textbf{Other related works.} The topic of unlearning has recently gained significant attention. Catastrophic forgetting (CF) \cite{Aleixo2023CatastrophicFI} is based on drastically forgetting previous data of neural networks upon learning new information. It proposes to continue training the model on the retain set until it forgets the forget set. Random Labels (RL) \cite{Hayase2020SelectiveFO} proposed using the entire retain and forget set, however, they randomly change the label of forget get to make it noisy then the model treats it as unimportant data. Inspired by adversarial attack, Boundary Shrink (BS) \cite{Chen2023BoundaryUR} proposed only using forget set to unlearn, by adopting the FGSM attack to assign incorrect labels to forget data points. Meanwhile, SALUN \cite{Fan2023SalUnEM} proposed a weight saliency-based approach to enhance MU performance and SCRUB \cite{Kurmanji2023TowardsUM} proposed a teacher-student unlearning algorithm running on both retain and forget set. 

However, compared to exact unlearning, approximated unlearning only weakly guarantees removing entire forget data. While aforementioned  unlearning methods only focus on accuracy on retain-forget set, and unlearning time complexity as well, privacy metrics are not considered in depth. Meanwhile, membership inference attack (MIA) success rate has been adopted as an evaluation metric \cite{Chen2023BoundaryUR, Fan2023SalUnEM, Kurmanji2023TowardsUM}, yet its real-world application is limited. In this paper, we raise a concern about the fairness and robustness properties of the unlearned model, which we believe is critical along with accuracy.

\section{Our Conjectures}

\textbf{Notation.}
We consider a classification problem of a neural network $f_{\theta_{\mathcal{D}}}$, with parameter $\theta_{\mathcal{D}}$ which is well-trained on the dataset $\mathcal{D}=\cup_{c=1}^C \mathcal{D}_c$, where $\mathcal{D}_c=\{x_i, c\}^N$, $C$ is the number of classes, and each class set $\{\mathcal{D}_c\}^C$ is independent of each other. The unlearning task is to remove the influence of the forget class $\mathcal{D}_f, f\in\{1,...,C\}$ from the model $f_{\theta_0}$ while ensuring the model performs well on the retain classes $\mathcal{D}_r=\mathcal{D} \setminus \mathcal{D}_f$. Let $\theta_{\mathcal{D}}$ and $\theta_r$ represent the parameters that are well-trained on datasets $\mathcal{D}$ and $\mathcal{D}_r$, respectively. And, the parameters $\theta_u = \mathcal{A}(\theta_{\mathcal{D}}, \mathcal{D}_f, \mathcal{D}_r)$ denote the unlearned parameters derived by the unlearning algorithm $\mathcal{A}$.

Assuming the model $f_\theta$ contains $L$ normalization layers, our work investigates the distribution of normalized feature vectors derived from normalization layers. In particular, we focus on its variance to establish a well-trained property. And, we denote $\sigma_{\mathcal{D}}^{l}$ and $\sigma_{c}^{l}$ as the variance from the $l^{th}$ normalization layer on the $\mathcal{D}$ set and $\mathcal{D}_c$, respectively.

 \textbf{Fairness-variance bound.}
The normalization layers normalize the feature distribution to a unit Gaussian distribution ($\sigma_\mathcal{D}=1)$, hence imposing an implicit constraint on the variance of each class. Given the assumption of independence among classes, we can use their variance to assess the differential treatment of each class in the model, referred to as the \textit{fairness property}. Initially, we represent the feature list in the $l_{th}$ layer of a dataset $\mathcal{D}_c$ as $\{f^l_i\}_{\mathcal{D}_c}, f_i^l\in R^{d^l}$. Assuming independence among each dimension in the feature vector, we define its variance as follows:

\begin{equation}
    \sigma^l_c = \frac{1}{d^l}\text{trace}\biggl(\Sigma\biggl[\biggl\{f^l_i\biggl\}_{\mathcal{D}_c}\biggl]\biggl),
\end{equation}
where $\Sigma$.[$\cdot$] is a covariance matrix of a set.

Consequently, we define a \textit{fairness-gap}, which represents the upper bound for the differentiation between variance among training classes as follows:

\begin{definition}[Fairness-gap]
  In classification setting, a fairness-gap of a neural network $f_\theta$ on $\mathcal{D}=\cup_{c=1}^C \mathcal{D}_c$ at $l_{th}$ normalization layer is
  \begin{equation}
   \epsilon^l   \coloneq  \max \{\sigma_c^l\}^C - \min \{\sigma_c^l\}^C. 
  \end{equation}
\end{definition}

Naturally, we desire our trained or fine-tuned models to perform effectively in real-world situations. The model must exhibit robustness to diverse form of attacks, including adversarial attacks. Hence, we define the robust learning algorithm as follows:

\begin{definition}[The robust learning algorithm]
  A robust learning/fine-tuning algorithm produces robust models.
\end{definition}

Although fairness is a crucial property, we expect that any fine-tuning methods, particularly unlearning in this scenario, should preserve model's fairness. We define a fine-tuning method that breaks the fairness balance makes the model vulnerable.

\begin{Conjecture}
  A robust machine unlearning algorithm should preserve the fairness-gap on $\mathcal{D}_r$ of the original model.
\end{Conjecture}

 \textbf{Fairness-robustness relationship.}
The variance-bias trade-off indicates that high variance implies overfitting, wherein the model is exceptionally sensitive to minor input modifications, such as those introduced by adversarial attacks. Furthermore, as previously mentioned, normalization layers impose a variance constraint for each class. Therefore, a higher fairness-gap implies the existence of high-variance classes, which means the model is more vulnerable to adversarial attack.

\begin{Conjecture}
  The higher fairness-gap is, the more the model is sensitive and vulnerable.
\end{Conjecture}

This Conjecture indicates that we can enhance the model's robustness by optimizing the fairness-gap between classes, such as by introducing an additional loss function to minimize the fairness-gap between each class pair. The correlation between fairness and robustness will be demonstrated in Section 4.

\begin{figure*}
\centering
\begin{subfigure}[t]{0.31\linewidth}
    \centering
    \includegraphics[scale=0.9]{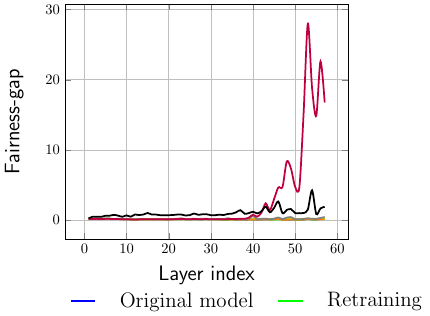}
    \caption{\large{ResNet50 (all models)}}
\end{subfigure}
\begin{subfigure}[t]{0.31\linewidth}
    \centering
    \hspace{3.65mm}
    \includegraphics[scale=0.9]{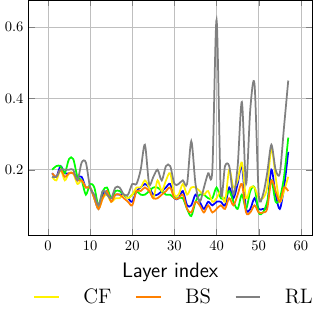}
    \caption{\large{ResNet50 (except SALUN and SCRUB)}}
 \end{subfigure}
\hfil
 \begin{subfigure}[t]{0.31\linewidth}
    \centering
    \hspace{-3.65mm}
     \includegraphics[scale=0.9]{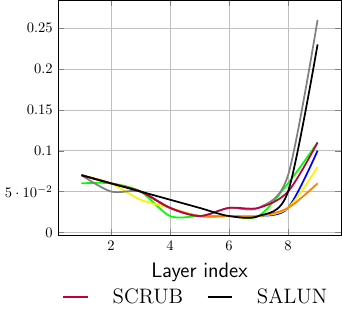}
    \caption{\large{SmallViT}}
\end{subfigure}
 \caption{The fairness-gap values in each normalization layer of ResNet50 and SmallViT on the retain-test dataset. In the ResNet model, the retraining model, CF, and BS show similar fairness-gaps to the original model, whereas RL, SALUN, and SCRUB demonstrate unstable fairness-gaps. In the ViT model, RL and SALUN show unstable behavior, whereas the others are close to the original model. The fairness-gap of ViT indicates greater stability than that of ResNet in unlearning.}
 \label{f1}
\end{figure*}

\section{Experiments}
\subsection{Experimental Setup}

\textbf{Dataset}. We conduct experiments on the \textsc{Cifar-10} dataset \cite{krizhevsky2009cifar}. Initially, we train a model on entire classes, referred to as the ``original model''. In the context of the unlearning task, we define the class ``trucks'' as the forget set, whereas all other classes are the retain set. Subsequently, we train a ``retraining model'' on the retain set. The retraining model is regarded as an optimal solution for the unlearning algorithms.

\textbf{Models}. This study involves experiments on ResNet50 \cite{He2015DeepRL} and SmallViT \cite{DosoViTskiy2020AnII}. We train ResNet50 from scratch for 400 epochs using Adam optimizer with a fixed learning rate of 0.0001, momentum of 0.9, weight decay of 0.0005, and batch size of 128. For ViT, we apply the same configuration used for ResNet training but extend the number of epoch to 1500 epochs.

\textbf{Unlearning setup}. We finetune the original model in 50 epochs by the following unlearning algorithms:
\begin{itemize}
    \item Catastrophic forgetting (CF) \cite{Aleixo2023CatastrophicFI}: Continue training the model on the retain set using the same configuration as during initial training.
    \item Random Labels (RL) \cite{Hayase2020SelectiveFO}: Fine-tune using both the retain and forget sets with the same configuration as training, randomly assigning a new label to the forget set in each iteration.
    \item Boundary Shrink (BS) \cite{Chen2023BoundaryUR}: Using only the forget set, we apply FGSM with a magnitude of 0.1 to create adversarially incorrect labels, using the SGD optimizer with a learning rate of 0.0001.
    \item SALUN \cite{Fan2023SalUnEM}: Using both the retain and forget sets, referred to as the unlearning setting from the original SALUN paper \cite{Fan2023SalUnEM}, we train it using SGD with a learning rate of 0.0001, a momentum of 0.9, and a weight decay of 0.0005.
    \item SCRUB \cite{Kurmanji2023TowardsUM}: Using both retain and forget sets, referred to as the unlearning setting from the original SCRUB paper \cite{Kurmanji2023TowardsUM}, we train the model using Adam optimizer with a learning rate of 0.0001.
\end{itemize}

\textbf{Robustness evaluation}. 
 We use adversarial attacks on the \textsc{Cifar-10} test set to evaluate the robustness of the models. We implement FGSM with a magnitude of $\eta=0.001$ for the ResNet model and $\eta=0.01$ for the ViT model. Higher accuracy will be the evidence for higher robustness.

\subsection{Experimental Results}

\begin{table}
    \caption{Accuracy of each unlearning approach on ResNet50 and SmallViT.}
    \centering
    \begin{tabular}{l|cc|cc}
         \hline
         \multicolumn{5}{c}{ResNet50} \\
         \hline
         &  $\mathcal{D}_r^{train}$ ($\uparrow$)  & $\mathcal{D}_f^{train}$  ($\downarrow$) &  $\mathcal{D}_r^{test}$ ($\uparrow$)  & $\mathcal{D}_f^{test}$ ($\downarrow$) \\
         \hline
         \hline
        Retraining & 0.99 & 0.00 & 0.92 & 0.00 \\
         \hline
        CF & 0.98 & 0.00& 0.88 & 0.00 \\
         \hline
        BS \cite{Chen2023BoundaryUR} & 0.89 & 0.09 & 0.92 & 0.18\\
        \hline
        RL \cite{Hayase2020SelectiveFO} & 0.99 & 0.00 & 0.78 &  0.00 \\
         \hline
        SALUN \cite{Fan2023SalUnEM} & 0.94 & 0.00 & 0.82 & 0.01  \\
         \hline
        SCRUB \cite{Kurmanji2023TowardsUM} & 0.97 & 0.00 & 0.88 & 0.00 \\
         \hline
         \hline

         \hline
         \multicolumn{5}{c}{SmallViT} \\
         \hline
          &  $\mathcal{D}_r^{train}$ ($\uparrow$) & $\mathcal{D}_f^{train}$ ($\downarrow$)  &  $\mathcal{D}_r^{test}$ ($\uparrow$)  & $\mathcal{D}_f^{test}$ ($\downarrow$) \\
         \hline
         \hline
        Retraining  & 0.99 & 0.00 & 0.75 & 0.00 \\
         \hline
        CF & 0.99 & 0.01 & 0.73 & 0.01\\
         \hline
        BS \cite{Chen2023BoundaryUR} & 0.93 & 0.10 & 0.72 & 0.15  \\
        \hline
        RL \cite{Hayase2020SelectiveFO} & 0.98 & 0.00  &  0.73 & 0.00 \\
         \hline
        SALUN \cite{Fan2023SalUnEM} & 0.98 &  0.02 & 0.69 & 0.00 \\
         \hline
        SCRUB \cite{Kurmanji2023TowardsUM} & 0.99 & 0.38 & 0.75  &  0.22 \\
         \hline
         \hline
    \end{tabular}
    \label{tab:my_label1}
\end{table}

\textit{Unlearning results}. The results of the ResNet50 and SmallViT models are presented in Table \ref{tab:my_label1}. In comparison to the retraining model, CF attains comparable accuracy on the forget set, with a little decrease on the retain set. Although BS exclusively utilizes the forget set, it cannot ensure accuracy on the retain set, resulting in lower results compared to other methods. Also, the accuracy on the test-forget set remains high for ResNet, and fails to reach 0\% for ViT. For RL, while it attains a perfect 0\% on the forget set, it results in a decrease on the retain set. In the case of SALUN, it is noted that their performance is high on the training set, though poor on the test set. At last, SCRUB demonstrates strong performance on ResNet; however, it cannot attain 0\% on the forget set on ViT.

 \subsubsection{Conjecture 1. Fairness-gap.} To verify Conjecture 1, we illustrate the fairness-gap of each normalization layer in both models in Figure \ref{f1}. In comparison to the original and retraining models, they indicate significant similarity. When treating the retraining model as a target for unlearning, our Conjecture 1 can be used as a criterion for unlearning algorithms, requiring a similar fairness-gap from the original model. In addition to eliminating the impact of the forget set and preserving high accuracy on retain data, the unlearning algorithm must preserve the model's robustness property.

In the ResNet model, the retraining model, CF, and BS preserve the fairness-gap, though RL exhibited minor fluctuations in the intermediate and last layers, and SALUN and SCRUB show substantial fluctuations in the last layers, ultimately breaking the fairness-gap upper bound established by the original model.

The ViT model has more stability than ResNet, as the fairness-gap of seven unlearned ViT models is below 0.3. The original model indicates a fairness-gap value of less than 0.1, however the retraining model, CF, BS, and SCRUB demonstrate similar values. In contrast, RL and SALUN show stronger instability, with their biggest gaps estimated at approximately 0.25. The fairness-gap values indicate instability in the intermediate and last layers. It proposes that fine-tuning only at the intermediate and last layers may suffice, hence enhancing efficiency in terms of time and memory complexity.

\subsubsection{Conjecture 2. Fairness - Robustness.}

Table \ref{tab:my_label3} presents the accuracy of each model under an adversarial attack. All approximated unlearning methods make models more vulnerable to adversarial noise. In the ResNet model, we have demonstrated that RL, SALUN, and SCRUB show a significant fairness-gap in previous experiments; consistently, their adversarial accuracy is the lowest compared to the retraining model, CF, and BS. Only the CF performance surprises us with low accuracy, as the fairness-gap is comparable to the retraining model. In the ViT model, the lower fairness-gap compared to ResNet results in less degradation in performance. In relation to the fairness-gap, both RL and SALUN show the most significant fairness-gap and the lowest adversarial accuracy. Our observation indicates that, though the proposed fairness-gap and adversarial performance lack perfect consistency, there are indications that they correlate with the model's robustness.

\begin{table}[H]
    \caption{Accuracy of each unlearning approach by adversarial attack. While retraining model achieve the best robustness, other unlearning methods make models more vulnerable to adversarial attacks.}
    \centering
    \resizebox{1\columnwidth}{!}{\begin{tabular}{l|c|c|c|c|c|c}
         \hline
        & \textbf{Retraining}  & \textbf{CF} & \textbf{BS}  & \textbf{RL} & \textbf{SALUN}  & \textbf{SCRUB} \\
         \hline \hline
       ResNet50 &  \textcolor{blue}{0.52} & \textcolor{red}{0.25} &  0.45 & \textcolor{red}{0.22} & \textcolor{red}{0.12}  & \textcolor{red}{0.19} \\
         \hline
       ViT &  \textcolor{blue}{0.63}  & 0.60 & \textcolor{red}{0.58}  & \textcolor{red}{0.58} & \textcolor{red}{0.56} & 0.62 \\
         \hline
    \end{tabular}%
    }
    \label{tab:my_label3}
\end{table}

\section{Limitations and Future Works}

While the connection between fairness-gap and the robustness of models has been shown on the \textsc{Cifar10} dataset, it would be more convincing to validate this on larger datasets such as \textsc{Cifar100} \cite{krizhevsky2009cifar}, \textsc{Tiny-Imagenet} \cite{Le2015TinyIV}, and \textsc{Imagenet} \cite{Deng2009ImageNetAL}. Furthermore, our studies only focus on classification problems, yet unlearning in Large Language Models (LLMs) \cite{Yao2023LargeLM, Yao2024MachineUO, Gundavarapu2024MachineUI} or Vision Language Models (VLMs) \cite{Li2024SingleIU, Zhou2024VisualIL, Hong_Lee_Woo_2024} is crucial because of their wide applicability. Consequently, it has the potential to do more in-depth study in these areas and address real-world issues. Also, our work emphasizes the strong need to propose a robust unlearning method that mitigates the underlying fairness-gap and enhances the safety and robustness of unlearned models.

\section{Conclusions}

The main objective of our work is to address concerns on fairness and robustness in machine unlearning, which is unexplored. By defining a fairness-gap, we provide two new Conjectures concerning the model's fairness and robustness, suggesting that a higher fairness-gap weakens the model's robustness to adversarial attacks. Our hypotheses are corroborated by the variance property and the experimental results on ResNet and ViT. We show that robustness evaluation can be utilized as an unlearning metric. For future directions, we suggest unlearning through the fine-tuning of the intermediate and last layers, which is sufficient and more efficient.

\begin{acks}
    This work was partly supported by Institute for Information \& communication Technology Planning \& evaluation (IITP) grants funded by the Korean government MSIT: (RS-2022-II221199, RS-2024-00337703, RS-2022-II220688, RS-2019-II190421, RS-2023-00230\\337, RS-2024-00356293, RS-2022-II221045, RS-2021-II212068, and RS-2024-00437849). 
\end{acks}

\bibliographystyle{ACM-Reference-Format}
\balance
\bibliography{sample-sigconf}

\end{document}